# A New Concept for an Obstacle Avoidance System for the AUV "SLOCUM Glider" Operation under Ice

Mike Eichhorn, Member, IEEE
Institute for Ocean Technology
National Research Council Canada
Arctic Avenue, P.O. Box 12093
St. John's, Newfoundland A1B 3T5, Canada
phone + 01 709-772-7986; fax + 01 709-772-2462
e-mail: Mike.Eichhorn@nrc-cnrc.gc.ca

*Abstract*- **This paper presents a concept for a control system for an autonomous underwater vehicle under ice using a "SLOCUM" underwater glider. The project concept, the separate working tasks for the next one-and-a-half years and the first results will be presented. In this context the structure of the obstacle avoidance system and a simulator structure with a sensor and environment simulation as well as the interfaces to the glider hardware will be discussed. As a first result of the main research, a graph-based algorithm for the path planning in a time-varying environment (variable ocean field, moving obstacles) will be described.**

## I. INTRODUCTION

The concept presented in this article is a part of a joint research plan between the National Research Council Canada Institute for Ocean Technology and Memorial University of Newfoundland, to explore ocean currents, salinity, temperature and depth of the Jakobshavn fjord in Western Greenland. The plan is to explore the year round ice-covered fjord using a "SLOCUM" underwater glider, which is a particular type of autonomous underwater vehicle (AUV).

The currently available configuration of underwater gliders does not allow for operations in constrained, unknown highly dynamic environments without surface access for communication and navigation. The "SLOCUM" glider has a basic seafloor collision avoidance algorithm using a single beam altimeter implemented. However for the dynamic environment, the glider faces in Jakobshavn fjord, which includes large numbers of icebergs of unknown shape and size, the avoidance algorithms are inadequate. It is the topic of this paper to introduce a concept for an obstacle avoidance system, tailored towards this specific problem as well as towards the particular platform.

The design of the control system meets the hardware and software requirements of the vehicles (sensors, dynamic vehicle behaviour, computer unit, software architecture) and their working conditions (sea current, icebergs, intact layer of ice). As a result of the restricted navigation accuracy, the computer performance, the existing sensors for obstacle detection and the working conditions under ice, a two level concept for the obstacle avoidance is proposed (see Figure 1).

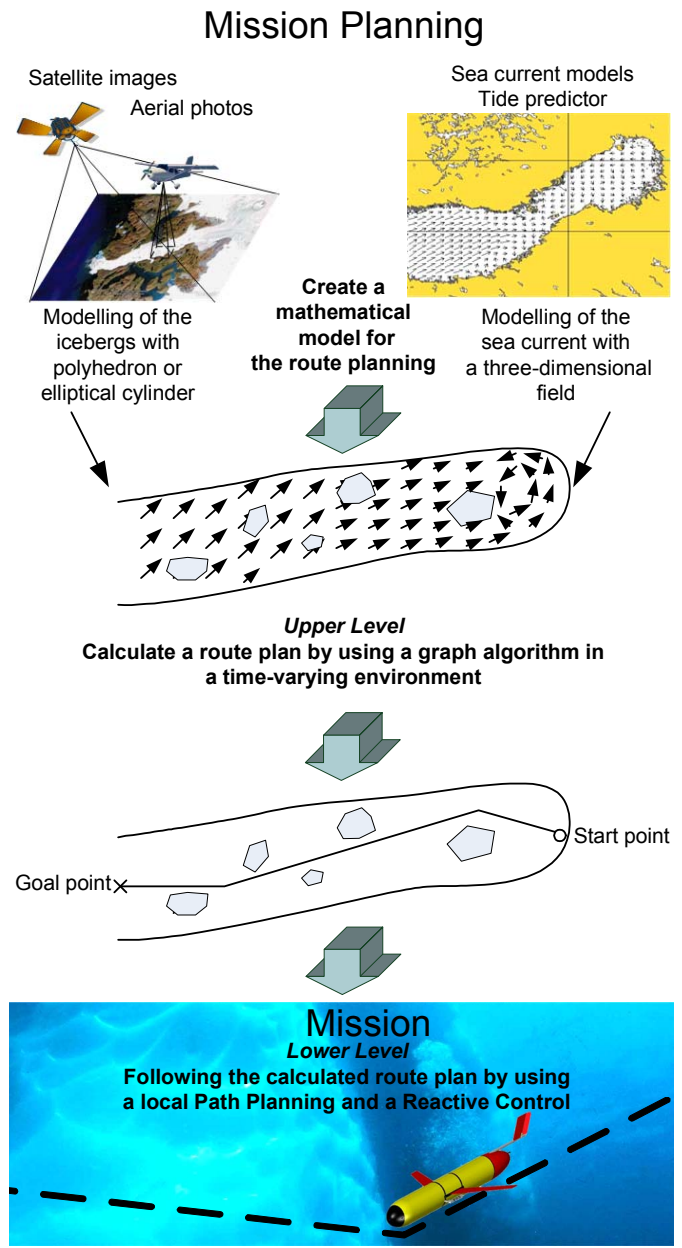

Figure 1. Two level concept of the Obstacle Avoidance System

The upper level generates a route plan offline by using the given operational area, the sea current information and the known positions of the icebergs from aerial and satellite photos. Through the restricted manoeuvrability of the gliders and the energy-optimized dive profile, the inclusion of the sea current information in the route planning is necessary. The sea current information can be generated from a tide predictor and a numerical simulation model. The generated plan of the route is sent at mission start to the vehicle.

In case of an unexpected or suddenly emerging obstacles as well as an error between the real and the approximated sea current during navigation along the generated route, the control will be given to the lower level of the obstacle avoidance system. There, depending on the computer availability, a local online *path planning* algorithm in combination with the *reactive control* or only the *reactive control* will be used. The local *path planning* algorithm uses information about the obstacles detected by the sonar and the sea current measured from the Acoustic Doppler Current Profiler (ADCP), to generate a local route plan. The *reactive control* reacts online to obstacles located in the proximity of the sonar by reactive course commands using the method of geometrical construction.

## II. MAIN RESEARCH AREAS

The working field for the development of an obstacle avoidance system for the "SLOCUM" glider contains the following four main research areas:

- Development of the algorithms for the route planning in a time-varying environment
- Design of a simulation test bed
- Development of the on-board obstacle avoidance system
- Implementation, testing and validation of the system

and the following interfaces to other research projects:

- Calculation of the time-varying current field using a current model for the offline mission planning
- Measure and detection of the sea current with an Acoustic Doppler Current Profiler during the mission [1, 2]
- Manoeuvring of the glider with low power propulsion for the reactive obstacle avoidance manoeuvres
- Modification of the existing glider controller software and building a new interface to the "SLOCUM" glider.

The requirements, the realization and first results of the route planning algorithm will be presented in section III. A short introduction to the other three working fields will be presented in the next sections.

### A. Simulation Test Bed

The simulation test bed contains a vehicle and an environment module. It is an extension to the existing glider simulator to test the new glider functionalities such as object detection, obstacle avoidance, manoeuvring with low power propulsion and the use of online sea current data (see Figure 2).

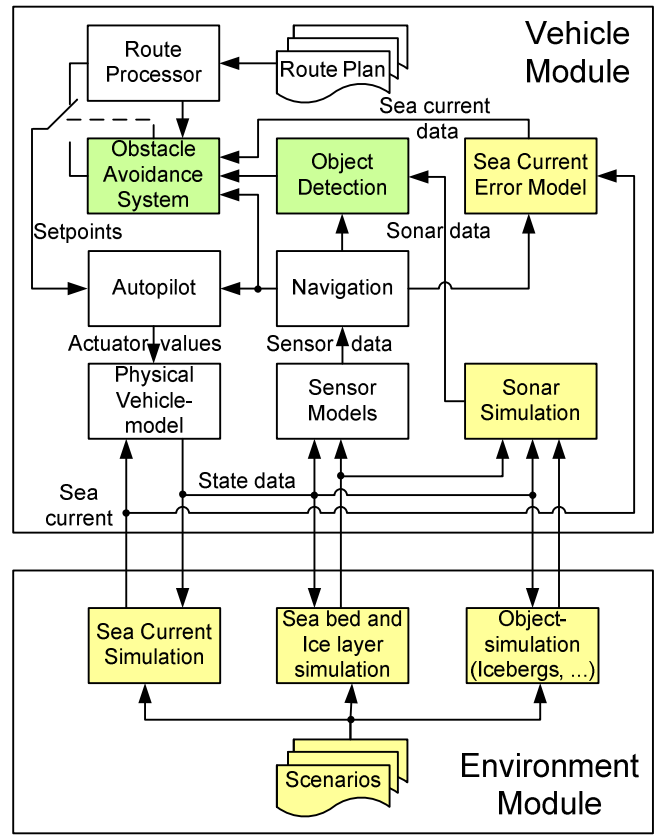

Figure 2. Components of the simulation test bed

The new modules will be written as object-oriented modules in C++ with defined interfaces to the other programs/modules. The environment module contains modules for the simulation of the sea bed, the ice layer as well as the icebergs and the sea current. The defined interfaces make it possible to use different complex models or to include existing models easily into the test bed.

An error model for the sea current data describes the errors associated with the measurements by the ADCP as well as the errors by the subsequent detection of the ADCP data in the module *Sea Current Detection* (see Figure 3). A separate consideration of the sensor and the detection algorithm is not necessary as the sea current detection module is tested separately. Furthermore the computing time of the error model corresponds to a fractional amount of the computing time of the complex simulation but provides the same predictions as without the error model.

The module *Sonar Simulation* generates sonar data which depends on the sensitivity and beam characteristics of the sonar and the attitude of the vehicle including all simulated geometrical objects in the environment module (sea bed, ice layer, icebergs).

This simulation test bed provided an easy checking of the interaction between the glider software and new algorithms with an appropriate mission plan in a simulated environment scenario.

## B. On-Board Obstacle Avoidance System

The development of the on-board obstacle avoidance system is a future portion in the second half of this project. This system will be designed under the conditions of the hardware and the software restrictions of the "SLOCUM" glider. Figure 3 shows the obstacle avoidance system with the several modules. This structure is based on a practical concept, which is used for the AUV "DeepC" and the AUV "MARIDAN 600" [3].

The *Collision Observation* module examines a collision possibility between the detected objects and the current way point list of the mission plan during the mission. If such a possibility is detected the *Evasive Module* will be activated. This module takes over the guidance of the vehicle and sends the control information to the Autopilot. In case of an evasive situation, the sub-module *Goal Point Generation* creates a point of rendezvous with the target route of the mission plan, using the actual plan of mission and the detected objects, which is the goal point for the obstacle avoidance system.

The *Evasive Module* is built in a two level structure (coloured grey in Figure 3). The upper level consists of *Path Planning*: a route avoiding the obstacles is generated here, using the existing information about the objects, the sea current and the environment. In the case of unexpected or suddenly emerging obstacles, during navigation on the generated evasive route, control will be passed to the lower level by the *Collision Observation* module. The *Reactive Control* level reacts to the located obstacles in the proximity of the sonar by reactive course commands. During the activation of the level *Reactive Control*, the route plan can be modified or recreated using the new object information in the level *Path Planning*. The first results from testing the algorithm for the *Path Planning* are described in section III. The *Reactive Control* uses a method of geometrical construction to create of gradient lines [4]. These lines represent a way to the goal point from every position in the area of operation.

An area of research in this part of the project is the analysis of the usability of the online generated sea current measured from the ADCP, to generate a local route plan as well as the choice of an applicable sonar system for the obstacle detection. Important requirements here are the restricted space inside the glider and low weight and energy consumption. Algorithms for the *Object Detection* module will be developed which exploit the object information from the sonar signals.

## C. Implementation, testing and validation of the system

The first sea trials will be begin in summer 2009 with the test of the offline path planning in the mission planning level. The test area is to the northeast offshore Newfoundland. Within this coastal area there exist accurate tide tables and sea current profiles which can be used for the design of the sea current model.

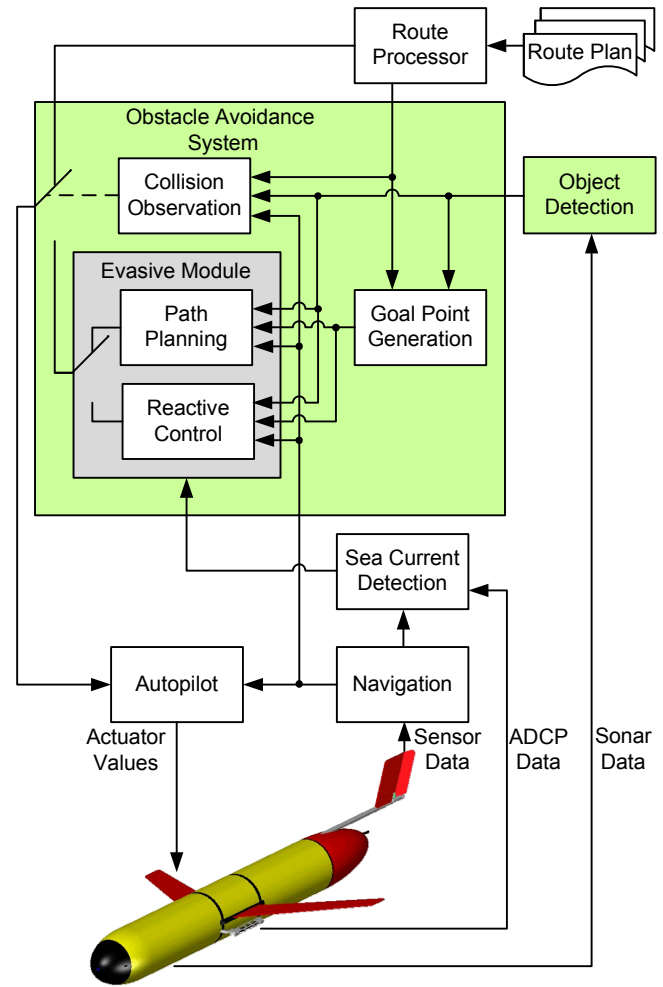

Figure 3. Structure of the Obstacle Avoidance System

Under deployment of this sea current model, a sea chart with a defined start-point and end-point, the path planning algorithm will calculate a mission route, which consists of a waypoint list for the existing operational system of the glider (GliderDOS [5]). The simulation test bed which is described in section A will be used for the software tests of the several modules of the Obstacle Avoidance System and their cooperation with other modules (see Figure 3).

The on-board *Obstacle Avoidance System* in combination with the *Object Detection* module will be tested in the tanks of the Institute for Ocean Technology during the winter 2009/2010. Spring 2010 will be the start of the sea tests of the on-board *Obstacle Avoidance System*. The first missions will test the online *Path Planning* using the sea current information from the ADCP. The offshore icebergs of Newfoundland represent the scenario for the tests of the *Object Detection* module with the on-board *Obstacle Avoidance System* in summer 2010. After completing all these tests, the Obstacle Avoidance System will be ready for a future glider mission in the Jakobshavn fjord in Western Greenland.

## III. PATH PLANNING FOR TIME-VARYING ENVIRONMENT

### A. Introduction

The task of the path planning algorithm used in an AUV is to find a time optimal path from a defined start position to a goal position by evasion of all static as well as dynamic obstacles in the area of operation with consideration of the dynamic vehicle behaviour and the time-varying ocean current flow. An additional constraint on the design of the algorithm is consideration of how the algorithm will be used on-board the "SLOCUM" glider in an online mode.

There exists a variety of solutions for the path planning in time-varying environment in the literature and especially for mobile autonomous systems. A generic algorithm was used for an AUV in [6] to find the path with minimum energy cost in a strong, time- and space-varying ocean current field. This approach finds a robust solution which will not necessarily correspond with the optimal solution. A symbolic wavefront expansion technique for an Unmanned Air Vehicle (UAV) in time-varying winds was introduced in [7]. The goal of this approach is to find the path and additionally the departure time for a minimum travel time between a start point and a goal point. All these approaches need a given, pre-defined mesh structure (see section B), unlike the following approaches.

A solution with a non-linear least squares optimization technique for a path planning of an AUV mission through the Hudson River was presented in [8]. The optimization parameters are a series of changeable nodes ($x_i$, $y_i$, $z_i$, $\Delta t_i$), which characterize the route. The inclusion of the time intervals $\Delta t_i$ allows a variation of the vehicle speed during the mission. In [9] a solution with optimal control to find the optimal trajectory for a glider in a time-varying ocean flow was presented. This approach applied the Nonlinear Trajectory Generation (NTG) algorithm including an ocean current flow B-spline model, a dynamic glider model as well a defined cost function which is a weighted sum of a temporal cost and an energy cost.

To solve all these requirements which were presented at the beginning of this section a search algorithm in a geometrical graph is preferred. This method makes it possible to find an optimal path by pre-defined criteria in a feasible computing time. The following sections present the algorithm and show the results with practical test scenarios.

### B. Generation of the geometrical graph

The geometrical graph is a mathematical model for the description of the area of operation with all its characteristics. Therefore points (vertices) within the operational area are defined which are passable by the vehicle. The passable connections between these points are recorded as edges in the graph. Every edge has a rating (cost, weight) which can be the length of the connection, the evolving costs for passing the connection or the time required for traversing the connection. There exist many approaches to describe an obstacle scenario with a low numbers of the vertices and the edges as possible, to decrease the computing time (visibility and quadtree graph [10]). In the case of the inclusion of an ocean current, the mesh structure of the graph will be decisive to its special change in gradient. In other words, the defined mesh structure should describe the trend of the ocean current flow in the operation area as effectively as possible. Therefore a uniform grid structure will be used. In the simplest case the edges are the connections between neighbouring obstacle free sectors (see Figure 4 a)). To achieve a shorter and smoother path for mobile robots additional edges to other sectors are implemented in [11] (see Figure 4 b)). The analyses of the found paths in a current field show that it is important to define a great number of edges with differences slopes (see Figure 4 c)). A further increase of the number of radiated edges leaves to increasing lengths which is not practicable to describe the change in gradient of the current flow.

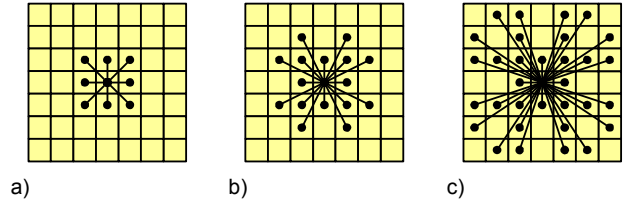

Figure 4. Grid structures a) 1-sector, b) 2-sector, c) 3-sector

### C. Graph-based Search-Algorithm

The described search method in this section is based on a classical Dijkstra algorithm [12, 13] which solves the single-source shortest-paths problem on a weighted directed graph. The exact solution by using a Dijkstra algorithm in a time-varying environment requires the inclusion of the time information as an additional dimension in the graph. For instance a 2D geometrical graph acquires additional layers for each defined discrete point of time. That will lead to very large graphs with many vertices (nodes) and edges as a result of using a fine time discretization.

The present algorithm uses the basic idea of the Dijkstra algorithm, to follow the shortest possible paths from a given source vertex and to avoid longer path segments during the updates of the neighbours. The algorithm searches for a path with the smallest travel time. This means the weight $w$ of the individual edges is the necessary time to travel along the path segment. This time depends on the length of the path segment and the speed $v_{path\_ef}$, with which the vehicle travels along the path in relation to the fixed world coordinate system. The speed $v_{path\_ef}$ depends on the vehicle speed through the water $v_{veh\_bf}$ (cruising speed), the amount and the direction of the ocean current vector as well as the direction of the path. The calculation of this speed and the travel time is described in [10, 14]. In the case that the sea current is so strong that the vehicle can no longer maintain the path, a large numerical value will be defined for the edge weight, and those segments will be excluded from the search. The generated mission plan guides the vehicle around those areas which would require a high degree of vehicle manoeuvrability to be able to negotiate its way through such areas.

The following table shows a comparison between the Dijkstra algorithm (left column) and the new algorithm TVE (time-varying environment) algorithm (right column). The syntax of the pseudo-code is adapted from [13]. The parameter G contains the graph structure with the vertex list and the edge list (*V* and *E*), $\pi$ is a predecessor list and *d* is the cost list for each vertex. *Q* is a priority queue that supports the EXTRACT-MIN and the DECREASE-KEY operations. The *color* list defines the current state of the vertex in the priority queue *Q*. (The allowable states are: WHITE: the vertex has not yet been discovered, GRAY: the vertex is in the priority list, BLACK: the vertex was checked). The gray marked text fields highlight the differences between the algorithms. There are the following three differences:

1. The new algorithm doesn't need the weight list *w* of the edges to begin the search. The algorithm needs a start time $t_o$ when the vehicle begins the mission.
2. The algorithm calculates the weight for the edge *w(u, v)* in a function *wfunc* during the search. This function calculates the travel time to drive along the edge from a start vertex *u* to an end vertex *v* by a given start time. The start time to be used will be the current cost value *d(u)*, which describes the travel time from the source vertex *s* to the vertex *u*.
3. A visited vertex *v* (*color[v]*=BLACK) can be checked several times and will be again included in the priority queue *Q* and will be available for further checking.

A detailed description of the algorithm, its restrictions and performance, possible modifications and extensions will be presented in [15].

TABLE I
PSEUDO-CODE OF THE ALGORITHMS

| DIJKSTRA(G, s, w) | TVE(G, s, $t_0$) |
|---|---|
| **for** each vertex $u \in V$ | **for** each vertex $u \in V$ |
|   $d[u] \leftarrow \infty$ |   $d[u] \leftarrow \infty$ |
|   $\pi[u] \leftarrow \infty$ |   $\pi[u] \leftarrow \infty$ |
|   color[u] ← WHITE |   color[u] ← WHITE |
| color[s] ← GRAY | color[s] ← GRAY |
| $d[s] \leftarrow 0$ | $d[s] \leftarrow t_0$ |
| INSERT(Q, s) | INSERT(Q, s) |
| **while** (Q≠Ø) | **while** (Q≠Ø) |
|   u ← EXTRACT-MIN(Q) |   u ← EXTRACT-MIN(Q) |
|   color[u] ← BLACK |   color[u] ← BLACK |
|   **for** each $v \in Adj[u]$ |   **for** each $v \in Adj[u]$ |
|     $d_v = w(u, v) + d[u]$ |     $d_v = wfunc(u, v, d[u]) + d[u]$ |
|     **if** ($d_v < d[v]$) |     **if** ($d_v < d[v]$) |
|       $d[v] \leftarrow d_v$ |       $d[v] \leftarrow d_v$ |
|       $\pi[v] \leftarrow u$ |       $\pi[v] \leftarrow u$ |
|       **if** (color[v] = GRAY) |       **if** (color[v] = GRAY) |
|         DEREASE-KEY(Q, v, $d_v$) |         DEREASE-KEY(Q, v, $d_v$) |
|       **else if** (color[v] = WHITE) |       **else** |
|         color[v] ← GRAY |         color[v] ← GRAY |
|         INSERT(Q, v) |         INSERT(Q, v) |
| **return** (d, $\pi$) | **return** (d, $\pi$) |

## D. Results

The following tests show the influence of the chosen grid structure of the predicted paths and the performance of the developed algorithm in a time-varying current field.

### 1. Stationary current field

The first test uses a time-invariant current field without obstacles. The task is to determine a time-optimal path when crossing a river, which possesses the current profile in equation (1). This profile shall describe the principle current flow in the Jakobshavn fjord created by tides.

$$v_{veh\_bf} = 2.2 \frac{m}{s}$$

$$v_{current,x} = 0 \quad v_{current,y} = \frac{4}{b^2} x(b-x) v_{current\_0} \quad (1)$$

with $b = 300m$, $v_{current\_0} = 1.8 \frac{m}{s}$

The following figures show the paths determined by the graph method and for comparison the time-optimal solution by optimal control. The geometrical graphs are produced with different grid sizes, aspect ratio and grid structures. TABLE II contains the travel times for various paths in a comparison with the best possible solution (optimal control) as well as the number of the vertices and edges of the appropriate geometrical graph.

A decrease of the grid size doesn't bring the desired improvement, as seen in Figure 5. A good choice of the aspect ratio is an important factor in determining the success of the path planning (see Figure 6). In the case of an aspect ratio of 1:2, the slope and the length of the diagonal and vertical edges in relation to the optimal route are so adverse that the predicted solution uses only the horizontal edge elements. The adding of new edges in the grid structure as presented in section III.B, leads to a good approximation of the predicted paths relative to the exact solution.

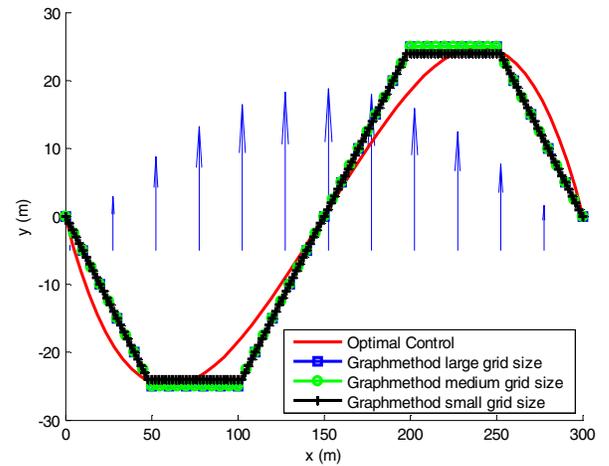

Figure 5. Path planning with the same aspect ratio and different grid sizes
(large: x=10m, y=5m; medium: x=5m y=2.5m; small x=2m, y=1m)

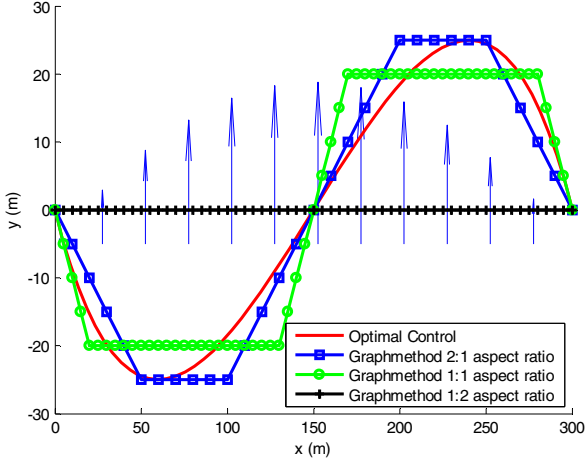

Figure 6. Path planning with different aspect ratios for grid structure 1-sector

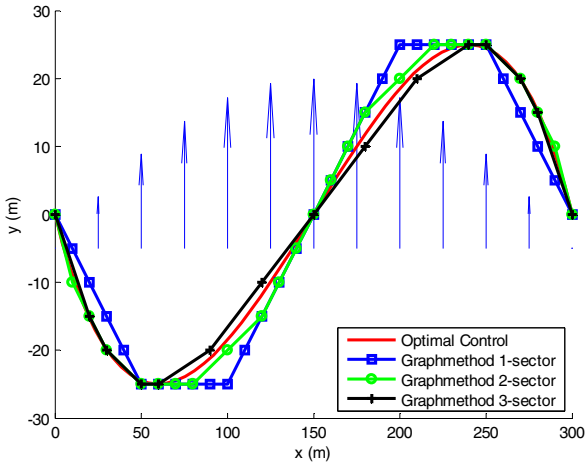

Figure 7. Path planning with different grid structures (see also Figure 4)

The adding of edges with new slopes gives new variation possibilities to the path search and leads to improvements in comparison to a grid structure with a lesser number of edges.

TABLE II
RESULTS OF THE DIFFERENT GRID STRUCTURES

| Method | time (s) | No. of Vertices | No. of Edges | Percentage time saving to Direct Drive | Percentage deviation to Optimal Control |
|---|---|---|---|---|---|
| Direct Drive | 167.52 | | | 0.0 | 10.96 |
| Optimal Control | 161.79 | | | 9.87 | 0.0 |
| Graph large | 165.48 | 403 | 2964 | 7.82 | 2.28 |
| Graph medium | 165.48 | 1525 | 11688 | 7.82 | 2.28 |
| Graph small | 165.47 | 9211 | 72420 | 7.83 | 2.27 |
| Graph 2:1 | 165.48 | 403 | 2964 | 7.82 | 2.28 |
| Graph 1:1 | 176.33 | 793 | 5904 | 1.77 | 8.99 |
| Graph 1:2 | 179.52 | 427 | 3012 | 0.0 | 10.96 |
| Graph 1-sector | 165.48 | 403 | 2964 | 7.82 | 2.28 |
| Graph 2-sector | 162.79 | 403 | 5676 | 9.32 | 0.62 |
| Graph 3-sector | 162.73 | 403 | 10612 | 9.56 | 0.35 |

### 2. Time-Varying Ocean Flow

The function used to represent a time-varying ocean flow describes a meandering jet in the eastward direction, which is a simple mathematical model of the Gulf Stream [16]. This function was applied in [6] to test the evolutionary path planning algorithm in a 2D ocean environment. The stream-function is:

$$\phi(x,y) = 1 - \tanh\left( \frac{y - B(t)\cos(k(x-ct))}{\left(1 + k^2 B(t)^2 \sin^2(k(x-ct))\right)^{\frac{1}{2}}} \right) \quad (2)$$

which uses a dimensionless function of an time-dependent oscillation of the meander amplitude

$$B(t) = B_0 + \varepsilon \cos(\omega t + \theta) \quad (3)$$

and the parameter set $B_0 = 1.2$, $\varepsilon = 0.3$, $\omega = 0.4$, $\theta = \pi/2$, $k = 0.84$ and $c = 0.12$ to describe the velocity field:

$$u(x,y,t) = -\frac{\partial \phi}{\partial y} \quad v(x,y,t) = \frac{\partial \phi}{\partial x} \quad (4)$$

The dimensionless value for the body fixed vehicle velocity $v_{veh\_bf}$ is 0.5. The exact solution was founded by solving a boundary value problem (BVP) with a collocation method [17] in MATLAB. The three ordinary differential equations (ODEs) include the two equations of motion:

$$\frac{dx}{dt} = u + v_{veh\_bf} \cos \varphi$$
$$\frac{dy}{dt} = v + v_{veh\_bf} \sin \varphi \quad (5)$$

and the optimal navigation formula from Zermelo [18]:

$$\frac{d\varphi}{dt} = -u_y \cos \varphi^2 + (u_x - v_y)\cos \varphi \sin \varphi + v_x \sin \varphi^2. \quad (6)$$

This approach is more robust for solving this time optimal problem than when trying to solve this same problem using a Hamilton equation.

Figure 8 shows the time sequence of the course through the time-varying ocean flow found by the TVE algorithm with a 3-sector grid structure. The determined solution has the characteristic that the vehicle drives in the mainstream of jet. This solution shows a very good agreement with the exact solution by optimal control in Figure 9. This figure contains the found paths by the TVE algorithm using different grid structures. The use of a 3-sector grid structure shows the best approximation with the exact solution but then this geometrical graph has the most vertices and edges. Here it is necessary to find a compromise between accuracy and the computing time. A further increase of the number of radiated edges (4…n-sector grid structure) leads to increasing lengths which is not practicable to describe the change in the gradient of the time-varying current flow. Figure 10 shows the predicted paths including stationary obstacles. A comparison of the results from all tests is presented in TABLE III.

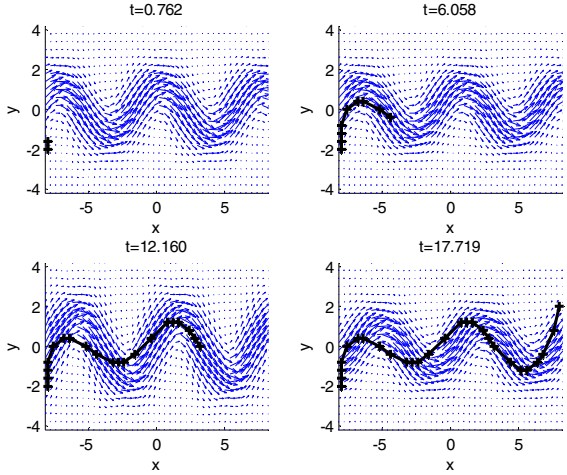

Figure 8. Time sequence of the predicted paths through the time-varying ocean current field by the TVE graph method with a 3-sector grid structure

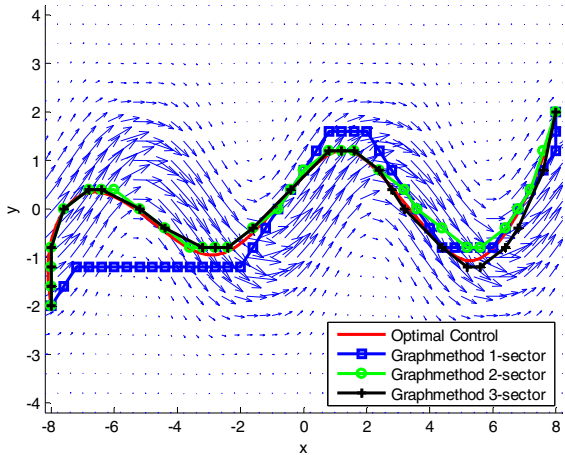

Figure 9. Time optimal paths through a time-varying ocean field using an exact numeric solution with Optimal Control and the TVE graph method

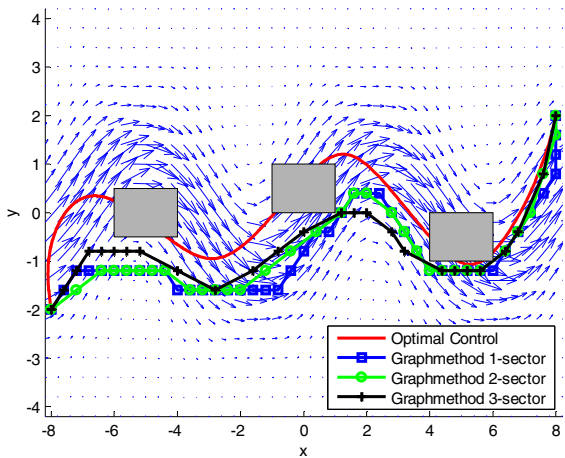

Figure 10. Time optimal paths through a time-varying ocean field with obstacles present

TABLE III
RESULTS OF THE DIFFERENT GRID STRUCTURES IN A TIME-VARYING CURRENT

| Method | time (s) | No. of Vertices | No. of Edges | Percentage deviation to Optimal Control |
|---|---|---|---|---|
| Direct Drive | ∞ | | | ∞ |
| Optimal Control | 17.195 | | | 0.0 |
| Graph 1-sector | 19.623 | 861 | 6520 | 14.12 |
| Graph 2-sector | 17.951 | 861 | 12680 | 4.40 |
| Graph 3-sector | 17.719 | 861 | 24296 | 3.05 |
| Graph 1-sector with obstacles | 23.094 | 816 | 5910 | 34.31 |
| Graph 2-sector with obstacles | 21.319 | 816 | 11195 | 23.99 |
| Graph 3-sector with obstacles | 20.903 | 816 | 20517 | 21.57 |

The two test scenarios used in this section show the performance of graph based methods by the search of a path through a complex sea current field. The accurateness of the predicted paths was compared to the paths predicted by optimal control. If the sea current is constant as in the first example a classical Dijkstra algorithm can also be used. If the time to travel along an edge is constant, the weights can be calculated before the search begins. However this approach doesn't perform well in by a time-varying environment. For this set of problems the new TVE algorithm was developed. The algorithm searches for the fastest path through a time-varying environment based on a given start time $t_0$. In this section only the sea current parameter is time-variant. By including moving obstacles in the scenario the function *wfunc* will be expanded by a collision calculation between the edge and the moving obstacles. This is a future research field. First results for the moving obstacles as well as a detailed description of the TVE algorithm, its restrictions and performance, possible modifications and extensions will be presented in [15].

It appears that the main problem associated with using a grid structure based search algorithm in a current field is the right choice of the grid size and the grid pattern. The use of expanded grid structures which were presented in section III.B leads to an improved approximation on the exact solution.

## IV. CONCLUSIONS

In this article a concept for an obstacle avoidance system for the AUV "SLOCUM Glider" operation under ice was presented. A possible application of the concept is a proposed research project at the Jakobshavn Glacier in Western Greenland. The first part of the paper gives an overview of the concept and the separate working tasks for the next one-and-a-half years. The use of the present simulation test bed allowed an easy and extensive checking of the designed algorithms together with the existing glider software in a defined test scenario. As a first result of the project a new graph-based search algorithm in a time-varying environment is presented in the second part of the paper. The influence of the grid structure will be determined by means of practical tests.

The first at-sea test of the path planning algorithm will begin in summer 2009. The goal of these tests is the validation of the offline generated mission plan with a real glider mission for the proposed test area to the northeast offshore Newfoundland [19].

ACKNOWLEDGMENT

This work is financed by the German Research Foundation (DFG) within the scope of a two-year research fellowship. I would like to thank the National Research Council Canada Institute for Ocean Technology in particular Dr. Christopher D. Williams for the support during this project. I gratefully acknowledge Dr. Ralf Bachmayer from the Memorial University of Newfoundland for the helpful discussions. Special thanks to Dr. Alberto Alvarez for his help with the sea current models.

REFERENCES

[1] NORTEK AS, "Aquadopp Profiler Installed on WEBB Glider," 2009, http://www.nortek-as.com/news/aquadopp-profiler-installed-on-webb-glider.
[2] B. Garau, A. Alvarez and G. Oliver, "AUV navigation through turbulent ocean environments supported by onboard H-ADCP," in *Proc. 2006 IEEE International Conference on Robotics and Automation*, pp. 3556-3561, Orlando Florida, USA, May 2006.
[3] M. Eichhorn, "Guidance of an Autonomous Underwater Vehicle in Special Situations," in *Proc. IEEE Oceans 2005 - Europe,* pp. 35-40, Brest, France, 20-23 June 2005.
[4] M. Eichhorn, "A Reactive Obstacle Avoidance System for an Autonomous Underwater Vehicle," in *Proc. 16th IFAC World Congress 2005*, Prague, Czech Republic, July 3-8 2005.
[5] Teledyne Webb Research, *User Manual, Slocum Glider,* 2008.
[6] A. Alvarez, A. Caiti and R. Onken, "Evolutionary Path Planning for Autonomous Underwater Vehicles in a Variable Ocean," *IEEE Journal of Oceanic Engineering,* vol. 29, no. 2, 2004, pp. 418-428.
[7] M. Soulignac, P. Taillibert, M. Rueher, "Path Planning for UAVs in Time-Varying Winds," in *Proc. The 27th Workshop of the UK PLANNING AND SCHEDULING Special Interest Group*, Heriot-Watt University, Edinburgh, United Kingdom, December 11-12, 2008.
[8] D. Kruger, R. Stolkin, A. Blum and J. Briganti, "Optimal AUV path planning for extended missions in complex, fast flowing estuarine environments," in *Proc. IEEE International Conference on Robotics and Automation*, Rom, Italy, April 10-14 2007.
[9] W. Zhang, T. Inanc, S. Ober-Blöbaum, J.E. Marsden, "Optimal Trajectory Generation for a Glider in Time-Varying 2D Ocean Flows B-spline Model," in *Proc. IEEE International Conference on Robotics and Automation*, pp. 1083-1088, Pasadena, CA, USA, May 19-23, 2008.
[10] M. Eichhorn, "An Obstacle Avoidance System for an Autonomous Underwater Vehicle," in *Proc. Proceedings of 2004 International Symposium on Underwater Technology*, pp. 75-82, 20.-23. April 2004
[11] T. Ersson and X. Hu, "Path Planning and Navigation of Mobile Robots in Unknown Environments," in *Proc. Intelligent Robots and Systems IEEE/RSJ International Conference*, pp. 858-864, Maui, HI, USA, 11/29-11/03 2001.
[12] E.W. Dijkstra, "A Note on Two Problems in Connexion with Graphs," *Numerische Mathematik,* no. 1, 1959, pp. 269-271.
[13] J.G. Siek, L. Lee and A. Lumsdaine, *The Boost Graph Library,* Addison-Wesley, 2002.
[14] R.E. Davis, N.E. Leonard and D.M. Fratantoni, "Routing strategies for underwater gliders," Elsevier, *Deep-Sea Research II,* 2008.
[15] Eichhorn, Mike, "Optimal Path Planning for AUVs in Time-Varying Ocean Flows," in *Proc. 16th Symposium on Unmanned Untethered Submersible Technology (UUST09)* , Durham NH, USA, August 23-26 2009, in press.
[16] M. Cencini, G. Lacorata, A. Vulpiani and E. Zambianchi, "Mixing in a Meandering Jet: A Markovian Approximation," *Journal of Physical Oceanography,* vol. 29, 1999, pp. 2578-2594.
[17] N. Hale and D.R. Moore, *A Sixth-Order Extension to the MATLAB Package bvp4c of J. Kierzenka and L. Shampine,* Oxford University Computing Laboratory, technical report, April 2008.
[18] E. Zermelo, "Über das Navigationsproblem bei ruhender oder veränderlicher Windverteilung," *Z. Angew. Math. Mech.,* vol. 11, no. 2, 1931, pp. 114-124.
[19] NCOG, "Newfoundland Center for Ocean Gliders," 2009, http://www.physics.mun.ca/~glider/.